\documentclass{article}

\usepackage{microtype}
\usepackage{amsmath}
\usepackage{graphicx}
\usepackage{subcaption}
\usepackage{booktabs} 
\usepackage{amssymb}
\usepackage{cancel}

\usepackage{hyperref}

\newcommand{\rulesep}{\unskip\ \vrule\ }

\newcommand{\fig}[1]{Fig.~\ref{fig:#1}}

\usepackage[accepted]{icml2018}

\icmltitlerunning{Stochastic Video Generation with a Learned Prior}

\begin{document}

\twocolumn[
\icmltitle{Stochastic Video Generation with a Learned Prior}



\begin{icmlauthorlist}
\icmlauthor{Emily Denton}{nyu}
\icmlauthor{Rob Fergus}{nyu,fb}
\end{icmlauthorlist}

\icmlaffiliation{nyu}{Courant Institute, New York University}
\icmlaffiliation{fb}{Facebook AI Research}

\icmlcorrespondingauthor{Emily Denton}{denton@cs.nyu.edu}

\icmlkeywords{Machine Learning, ICML}

\vskip 0.3in
]



\printAffiliationsAndNotice{}  

\begin{abstract}
  Generating video frames that accurately predict future world states
  is challenging. Existing approaches either fail to capture the full
  distribution of outcomes, or yield blurry generations, or both.  In
  this paper we introduce an unsupervised video generation model that learns a prior
  model of uncertainty in a given environment. Video frames are
  generated by drawing samples from this prior and combining them with
  a deterministic estimate of the future frame. The approach is
  simple and easily trained
  end-to-end on a variety of datasets. Sample generations are both
  varied and sharp, even many frames into the future, and compare
  favorably to those from existing approaches.

\end{abstract}

\section{Introduction}
\label{intro}

Learning to generate future frames of a video sequence is a challenging research problem with great relevance to reinforcement learning, planning and robotics. Although impressive generative models of still images have been demonstrated (e.g. \citet{reed17,nvidiagan}), these techniques do not extend to video sequences. The main issue is the inherent uncertainty in the dynamics of the world. For example, when a bouncing ball hits the ground unknown effects, such surface imperfections or ball spin, ensure that its future trajectory is inherently random. 

Consequently, pixel-level frame predictions of such an event degrade when a deterministic loss function is used, e.g.~with the ball itself blurring to accommodate multiple possible futures. Recently, loss functions that impose a distribution instead have been explored. One such approach are adversarial losses \cite{goodfellow2014}, but training difficulties and mode collapse often mean the full distribution is not captured well. 

We propose a new stochastic video generation (SVG) model that combines a deterministic frame predictor with time-dependent stochastic latent variables. 
We propose two variants of our model: one with a fixed prior over the latent variables (SVG-FP) and another with a learned prior (SVG-LP).
The key insight we leverage for the learned-prior model is that for the majority of the ball's trajectory, a deterministic model suffices. Only at the point of contact does the modeling of uncertainty become important. 
The learned prior can can be interpreted as a {\em a predictive model of uncertainty}.
For most of the trajectory the prior will predict low uncertainty, making the frame estimates deterministic. However, at the instant the ball hits the ground it will predict a high variance event, causing frame samples to differ significantly. 

We train our model by introducing a recurrent inference network to estimate the latent distribution for each time step.
This novel recurrent inference architecture facilitates easy end-to-end training of SVG-FP and SVG-LP.
We evaluate SVG-FP and SVG-LP on two real world datasets and a stochastic variant of the Moving MNIST dataset.
Sample generations are both varied and sharp, even many frames into the future.



\section{Related work}
\label{related}

Several models have been proposed that use prediction within video to learn deep feature representations appropriate for high-level tasks such as object detection. 
\citet{wang2015unsupervised} learn an embedding for patches taken from object video tracks. \citet{Zou12} use similar principles to learn features that exhibit a range of complex invariances. 
\citet{lotter2016deep} propose a predictive coding model that learns features effective for recognition of synthetic faces, as well as predicting steering angles in the KITTI benchmark.
Criterions related to slow feature analysis \cite{Wiskott02} have been proposed such as linearity of representations \cite{goroshin2015} and equivariance to ego-motion \cite{jayaraman15}.
\citet{agrawal15} learn a representation by predicting transformations obtained by ego-motion.


A range of deep video generation models have recently been proposed. 
\citet{Srivastava15} use LSTMs trained on pre-learned low dimensional image representations.
\citet{Ranzato14} adopt a discrete vector quantization approach inspired by text models.
 Video Pixel Networks \cite{Kalchbrenner16} are a probabilistic approach to generation whereby pixels are generated one at a time in raster-scan order (similar autoregressive image models include \citet{salimans2017pixelcnn++, Oord16}). 
Our approach differs from these in that it uses continuous representations throughout and generates the new frame directly, rather than via a sequential process over scale or location. 

\citet{Finn16} use an LSTM framework to model motion via transformations of groups of pixels.  
Other works predict optical flows fields that can be used to
extrapolate motion beyond the current frame,
e.g. \cite{liu09,visualdynamics16,walker15}. Although we directly generate pixels, our model also computes local transformations but in an implicit fashion. Skip connections between encoder and decoder allow direct copying of the previous frame, allowing the rest of the model to focus on changes. However, our approach is able handle stochastic information in a more principled way.

Another group of approaches factorize the video into static and dynamic components before learning predictive models of the latter. 
\citet{denton17} decomposes frames into content and pose representations using a form of adversarial loss to give a clean separation. An LSTM is then applied to the pose vectors to generate future frames.    
\citet{Villegas17a} do pixel level prediction using an LSTM that separates out motion and content in video sequences. A reconstruction term is used that combines mean squared error and a GAN-based loss. Although our approach also factorizes the video, it does so into deterministic and stochastic components, rather than static/moving ones. This is an important distinction, since the difficulty in making accurate predictions stems not so much from the motion itself, but the uncertainty in that motion.

\citet{Villegas17b} propose a hierarchical model that first generates high level structure of a video and then generates pixels conditioned on the high level structure. This method is able to successfully generate complex scenes, but unlike our unsupervised approach, requires annotated pose information at training time.

\citet{Chiappa17} and \citet{Oh15} focus on action-conditional prediction in video game environments, where known actions at each frame are assumed. 
These models produce accurate long-range predictions. In contrast to the above works, we do not utilize any action information. 

Several video prediction approaches have been proposed that focus on handling the inherent uncertainty in predicting the future. 
\citet{Mathieu15} demonstrate that a loss based on GANs can produce sharper generations than traditional
$\ell_2$-based losses. \citet{Vondrick16} train a generative adversarial network that separates out foreground and background generation. 
\citet{Vondrick17} propose a model based on transforming pixels from the past and train with an adversarial loss. 
All these approaches use a GAN to handle uncertainty in pixel-space and this introduces associated difficulties with GANs, i.e. training instability and mode collapse. By contrast, our approach only relies on an $\ell_2$ loss for pixel-space reconstruction, having no GANs or other adversarial terms.

Other approaches address uncertainty in predicting the future by introducing latent variables into the prediction model. 
\citet{henaff17} disentangle deterministic and stochastic components of a video by encoding prediction errors of a deterministic model in a low dimensional latent variable. 
This approach is broadly similar to ours, but differs in the way the latent variables are inferred during training and sampled at test time.
The closest work to ours is that of \citet{Babaeizadeh1017}, who  propose a variational approach from which stochastic videos can be sampled. We discuss the relationship to our model in more depth in Section \ref{sec:rw_finn}. 

Stochastic temporal models have also been explored outside the domain of video generation.
\citet{bayer2014} introduce stochastic latent variables into a recurrent network in order to model music and motion capture data. This method utilizes a recurrent inference network similar to our approach and the same time-independent Gaussian prior as our fixed-prior model. 
Several additional works train stochastic recurrent neural networks  to model speech, handwriting, natural language \cite{chung2015, fraccaro2016, bowman2016}, perform counterfactual inference \cite{krishnan2015} and anomaly detection \cite{Slch2016VariationalIF}. 
As in our work, these methods all optimize a bound on the data likelihood using an approximate inference network. They differ primarily in the parameterization of the approximate posterior and the choice of prior model.

\section{Approach}
\label{approach}

\newcommand{\x}{\textbf{x}}
\newcommand{\z}{\textbf{z}}

We start by explaining how our model generates new video frames,
before detailing the training procedure. Our model has two distinct
components: (i) a prediction model $p_\theta$ that generates the next
frame $\hat\x_t$, based on previous ones in the sequence $\x_{1:t-1}$
and a latent variable $\z_t$ and (ii) a prior distribution $p(\z)$ from which $\z_t$ is sampled at {\em at each time step} . The prior distribution can
be fixed (SVG-FP) or learned (SVG-LP).  Intuitively, the latent
variable $\z_t$ carries all the stochastic information about the next
frame that the deterministic prediction model cannot capture. After conditioning
on a short series of real frames, the model can generate multiple
frames into the future by passing generated frames back into the input
of the prediction model and, in the case of the SVG-LP model, the
prior also.

\begin{figure}[t!]
    \centering
  \includegraphics[width=\linewidth]{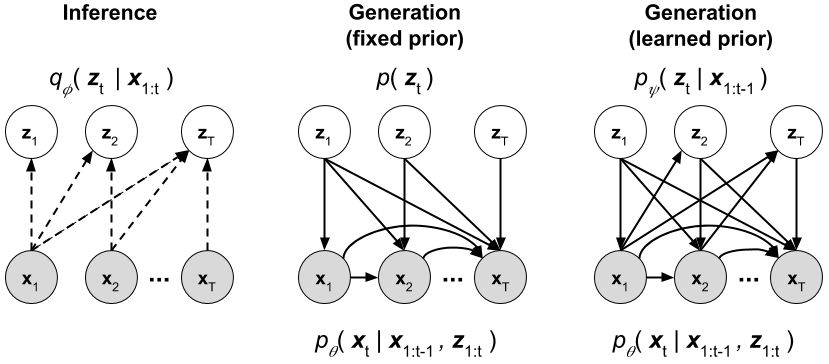}
 \caption{Inference (left) and generation in the SVG-FP (middle) and SVG-LP models (right).}
    \label{fig:gmodel}
\end{figure}

The model is trained with the aid of a separate inference model (not
used a test time). This takes as input the frame $\x_t$, i.e.~the
target of the prediction model, and previous frames $\x_{1:t-1}$. From this it computes a distribution
$q_\phi(\z_t|\x_{1:t})$ from which we sample $\z_t$. To prevent $\z_t$ just copying
$\x_t$, we force $q_\phi(\z_t|\x_{1:t})$ to be close to the prior distribution
$p(\z)$ using a KL-divergence term. This constrains the information
that $\z_t$ can carry, forcing it to capture new information not present in previous frames. A second term in the loss
penalizes the reconstruction error between $\hat\x_t$ and
$\x_t$. 
\fig{gmodel}a shows the inference procedure for both SVG-FP and SVG-LP. 
The generation procedure for SVG-FP and SVG-LP are shown in \fig{gmodel}b and \fig{gmodel}c respectively.

To further explain our
model we adopt the formalism of variational auto-encoders. 
Our recurrent frame predictor $p_{\theta}(\x_t | \x_{1:t-1}, \z_{1:t})$ is specified by a fixed-variance conditional Gaussian distribution
$\mathcal{N}(\mu_{\theta}(\x_{1:t-1},\z_{1:t}), \sigma)$. 
In practice, we set $\hat{\x_t} = \mu_{\theta}(\x_{1:t-1},\z_{1:t})$, i.e. the mean of the
distribution, rather than sampling. 
Note that at time step $t$ the frame predictor only receives $\x_{t-1}$ and $\z_t$ as input. 
The dependencies on all previous $\x_{1:t-2}$ and $\z_{1:t-1}$ stem from the recurrent nature of the model. 

Since the true distribution over latent variables $\z_t$ is intractable
we rely on a time-dependent inference network $q_{\phi}(\z_t |
\x_{1:t})$ that approximates it with a conditional Gaussian
distribution $\mathcal{N}(\mu_{\phi}(\x_{1:t}),
\sigma_{\phi}(\x_{1:t}))$. 
The model is trained by optimizing the variational lower bound: 
\vspace{-1mm}
\begin{align*}
 \mathcal{L}_{\theta,\phi}(\x_{1:T}) = \sum_{t=1}^T \big{[} &\mathbb{E}_{q_{\phi}(\z_{1:t} | \x_{1:t})} \log p_{\theta}(\x_t | \x_{1:t-1}, \z_{1:t}) \\
 - &\beta D_{KL}(q_{\phi}(\z_t | \x_{1:t}) || p(\z))\big{]}
\end{align*}
 
Given the form of $p_\theta$ the likelihood term reduces to an
$\ell_2$ penalty between  $\hat\x_t$ and $\x_t$. 
We train the model using the re-parameterization trick \cite{kingma2014a} and by estimating the expectation over $q_{\phi}(\z_{1:t} | \x_{1:t})$ with a single sample.
See Appendix \ref{bound} for a full derivation of the loss.

The hyper-parameter $\beta$ represents the trade-off between minimizing frame prediction error and fitting the prior.
A smaller $\beta$ increases the capacity of the inference network.
If $\beta$ is too small the inference network may learn to simply copy the target frame $\x_t$, resulting in low prediction error during training but poor performance at test time due to the mismatch between the posterior $q_{\phi}(\z_t | \x_{1:t})$ and the prior $p(\z_t)$.
If $\beta$ is too large, the model may under-utilize or completely ignore latent variables $\z_t$ and reduce to a deterministic predictor.
In practice, we found $\beta$ easy to tune, particularly for the learned-prior variant we discuss below. For a discussion of hyperparameter $\beta$ in the context of variational autoencoders see  \citet{higgins}.

\begin{figure*}[t!]
    \centering
\begin{subfigure}{0.28\textwidth}
  \includegraphics[width=\linewidth]{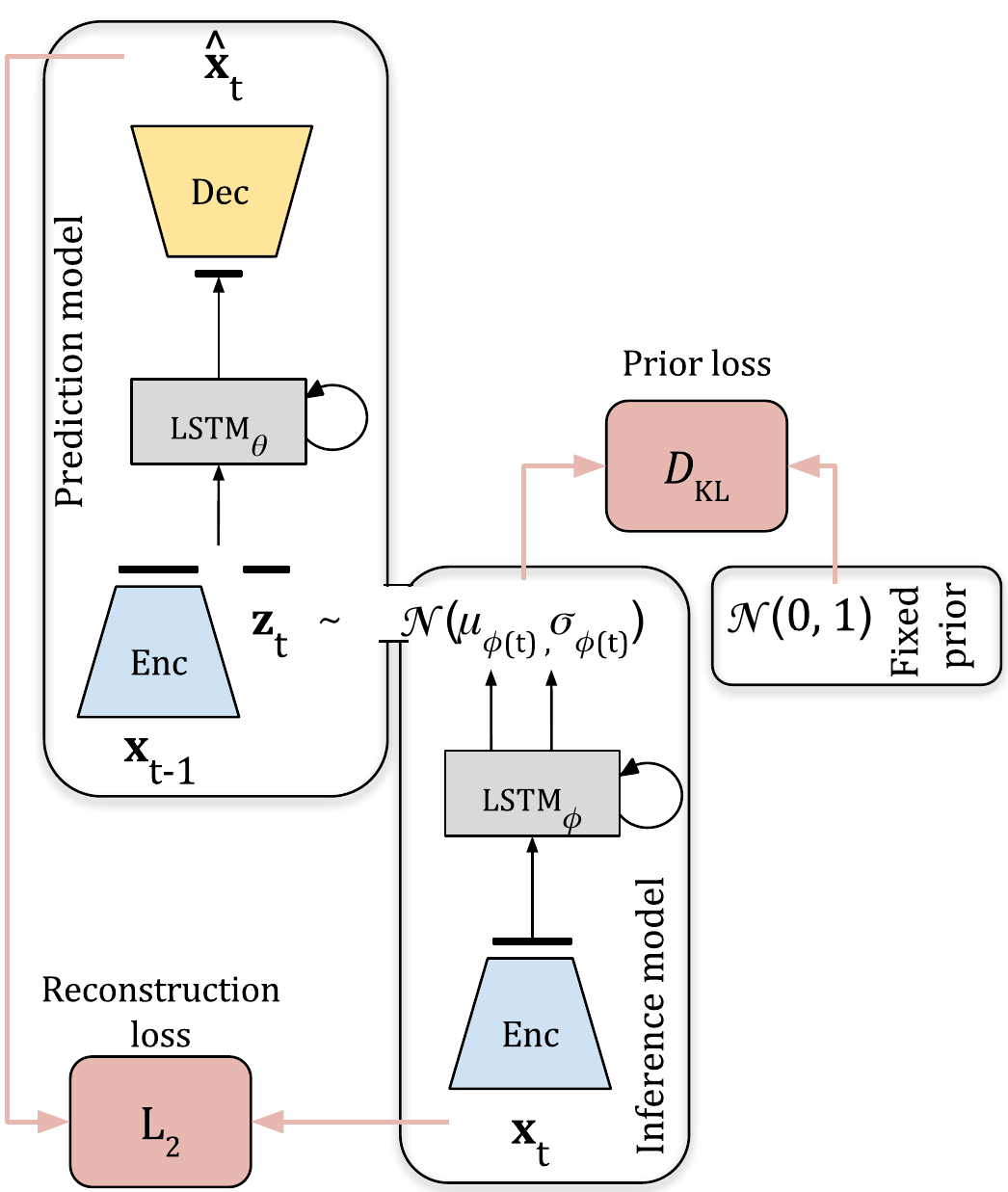}
\caption{}\label{fig:model_a}
\end{subfigure}
  \hspace{3mm}
  \rulesep
  \hspace{3mm}
\begin{subfigure}{0.28\textwidth}
  \includegraphics[width=\linewidth]{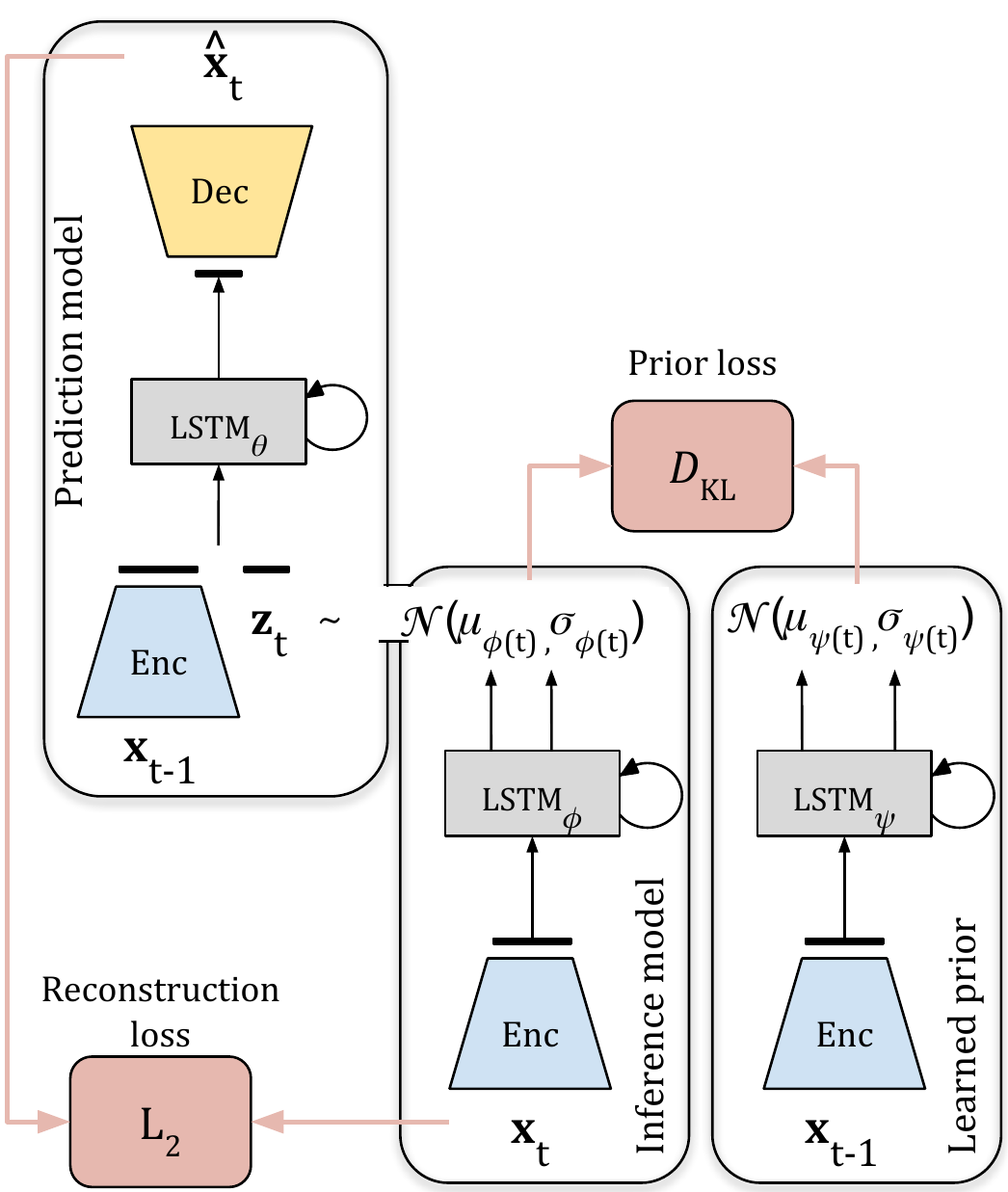}   
\caption{}\label{fig:model_b}
\end{subfigure}
  \hspace{3mm}
  \rulesep
  \hspace{3mm}
\begin{subfigure}{0.20\textwidth}
  \includegraphics[width=\linewidth]{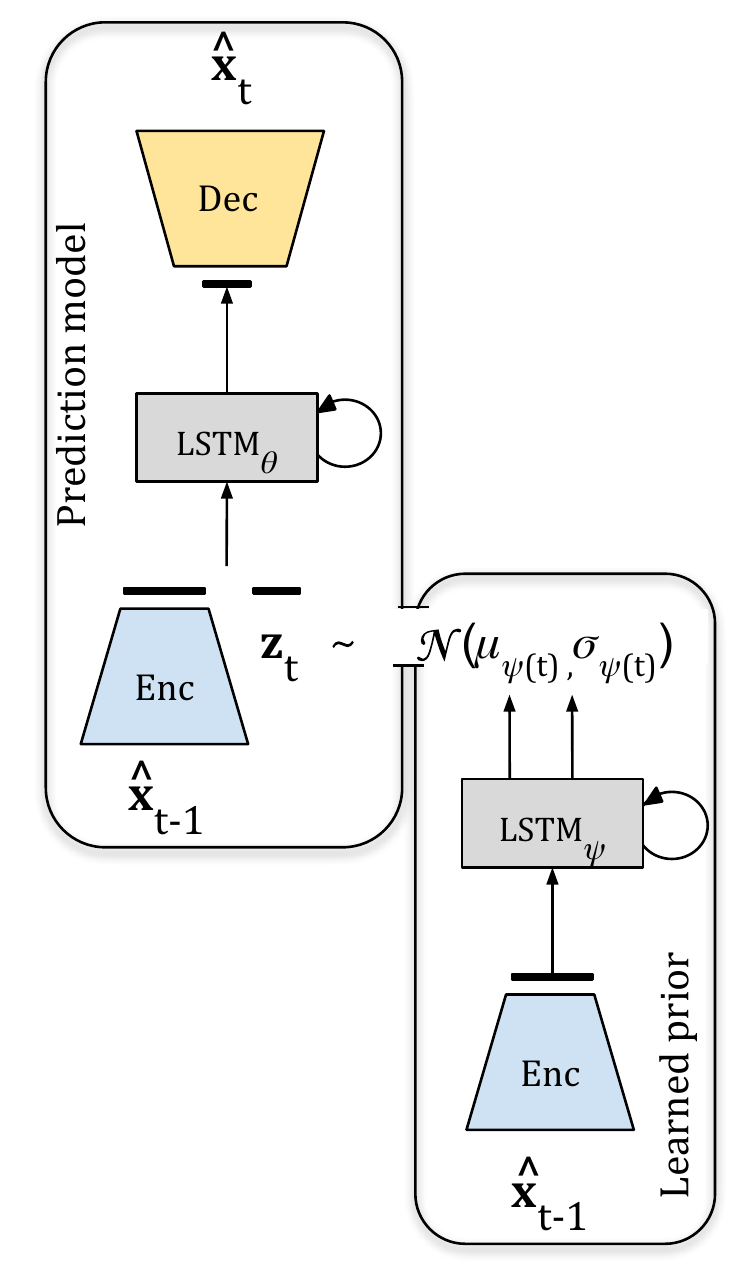}
\caption{}\label{fig:model_c}
\end{subfigure}
 \caption{Our proposed video generation model. (a) Training with a fixed prior (SVG-FP); (b) Training with learned prior (SVG-LP); (c) Generation with the learned prior model. The red boxes show the loss functions used during training. See text for details. }
    \label{fig:model}
\end{figure*}

{\bf Fixed prior:} The simplest choice for $p(\z_t)$ is a fixed Gaussian
$\mathcal{N}(0, \textbf{I})$, as is typically used in variational auto encoder
models. We refer to this as the SVG-FP model, as shown in
\fig{model_a}. A drawback is that samples at each time
step will be drawn randomly, thus ignore temporal dependencies present
between frames. 

{\bf Learned prior:} A more sophisticated approach is to learn a prior that varies
across time, being a function of all past frames up to {\em but not including} the frame being predicted $p_{\psi}(\z_t | \x_{1:t-1})$. 
Specifically, at time $t$ a prior network observes frames $\x_{1:t-1}$
and output the parameters of a conditional Gaussian distribution
$\mathcal{N}(\mu_{\psi}(\x_{1:t-1}), \sigma_{\psi}(\x_{1:t-1}))$. 
The prior network is trained jointly with the rest of the model by maximizing:
\vspace{-1mm}
\begin{align*}
 \mathcal{L}_{\theta,\phi,\psi}(\x_{1:T}) =  \sum_{t=1}^T \big{[} &\mathbb{E}_{q_{\phi}(\z_{1:t} | \x_{1:t})} \log p_{\theta}(\x_t | \x_{1:t-1}, \z_{1:t}) \\
 - &\beta D_{KL}(q_{\phi}(\z_t | \x_{1:t}) || p_{\psi}(\z_t | \x_{1:t-1}))\big{]}
\end{align*}
We refer to this model as SVG-LP and illustrate the training procedure in \fig{model_b}.

At test time, a frame at time $t$ is generated by first sampling $\z_t$ from the prior.
In SVG-FP we draw $\z_t \sim \mathcal{N}(0, \textbf{I})$ and in SVG-LP we draw $\z_t \sim p_{\psi}(\z_t | \x_{1:t-1})$.
Then, a frame is generated by $\hat{\x}_t = \mu_{\theta}(\x_{1:t-1}, \z_{1:t})$.
After conditioning on a short series of real frames, the model begins to pass generated frames $\hat{\x}_{t}$ back into the input of the prediction model and, in the case of the SVG-LP model, the prior.
The sampling procedure for SVG-LP is illustrated in \fig{model_c}.

{\bf Architectures}: We use a generic convolutional LSTM for
$p_{\theta}$, $q_{\phi}$ and $p_{\psi}$. Frames are input to the LSTMs
via a feed-forward convolutional network, shared across all
three parts of the model. A convolutional frame decoder maps the output of the frame predictor's recurrent network back to pixel space.

For a time step $t$ during training, the generation is as follows, where the LSTM recurrence is omitted for brevity:
\begin{align*}
\mu_{\phi}(t), \sigma_{\phi}(t) &= LSTM_{\phi}(h_t) \hspace{16mm} h_t = Enc(\x_t) \\
\z_t &\sim \mathcal{N}(\mu_{\phi}(t), \sigma_{\phi}(t)) \\
g_t &= LSTM_{\theta}(h_{t-1}, \z_t) \hspace{5mm} h_{t-1} = Enc(\x_{t-1}) \\
\mu_{\theta}(t) &= Dec(g_t)
\end{align*}

During training, the parameters of the encoder $Enc$ and decoder $Dec$
are also learned, along with the rest of the model, in an end-to-end
fashion (we omit their parameters from the loss functions above for
brevity). 

In the learned-prior model (SVG-LP), the parameters of the prior distribution at time $t$ are generated as follows, where the LSTM recurrence is omitted for brevity:
\begin{align*}
h_{t-1} &= Enc(\x_{t-1}) \\
\mu_{\psi}(t), \sigma_{\psi}(t) &= LSTM_{\psi}(h_{t-1}) \\
\end{align*}

\vspace{-4mm}
\subsection{Discussion of related models}
\label{sec:rw_finn}
Our model is related to a recent stochastic variational video prediction model of \citet{Babaeizadeh1017}.
Although their variational framework is broadly similar, a key
difference between this work and ours is the way in which the latent
variables $\z_t$ are estimated during training and sampled at test
time.

The inference network of \citet{Babaeizadeh1017} encodes the {\em entire
video sequence} via a feed forward convolutional network to estimate
$q_{\theta}(\z | \x_{1:T})$. They propose two different models that
use this distribution. In the time-invariant version, a single $\z$
is sampled for the entire video sequence. 
In the time-variant model, a different $\z_t \sim q_{\theta}(\z |
\x_{1:T})$ is sampled for every time step, all samples coming from the
{\em same distribution}.

In contrast, both our fixed-prior and learned-prior models utilize a
more flexible inference network that outputs a different posterior
distribution for every time step given by $q_{\theta}(\z_t |
\x_{1:t})$ (note $\x_{1:t}$, not $\x_{1:T}$ as above).

At test time, our fixed-prior model and the time-variant model of
\citet{Babaeizadeh1017} sample $\z_t$ from a fixed Gaussian prior at
every time step. By contrast, our learned-prior model draws samples
from the time-varying distribution: $p_{\psi}(\z_t | \x_{1:t-1})$,
whose parameters $\psi$ were estimated during training.

These differences manifest themselves in two ways. First, the
generated frames are significantly sharper with both our models (see
direct comparisons to \citet{Babaeizadeh1017}  in
Figure \fig{bair}). Second, training our model is much
easier. Despite the same {\em prior} distribution being used for both our fixed-prior model and \citet{Babaeizadeh1017}, the time variant {\em posterior} distribution introduced in our model appears crucial for successfully training the model.
Indeed, \citet{Babaeizadeh1017} report difficulties training their model by naively optimizing the variational lower bound, noting that the model simply ignores the latent variables. 
Instead, they propose a scheduled three phase training procedure whereby  first the deterministic element of the model is trained, then latent variables are introduced but the KL loss is turned off and in the final stage the model is trained with the full loss. 
In contrast, both our fixed-prior and learned-prior models are easily
trainable end-to-end in a single phase using a unified loss
function.

\section{Experiments}
\label{experiments}

\begin{figure*}[t!]
\centering
     \includegraphics[width=0.98\linewidth]{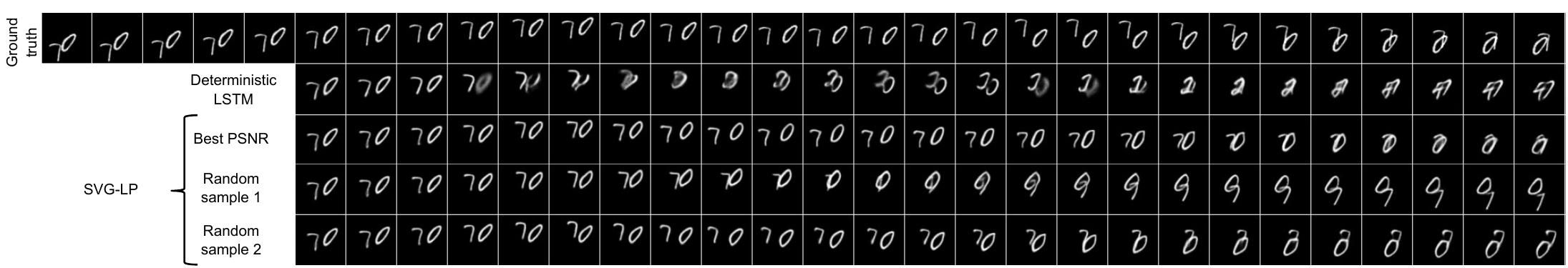}
 \caption{Qualitative comparison between SVG-LP and a
   purely deterministic baseline. The deterministic model
   produces sharp predictions until ones of the digits collides with a 
   wall, at which point the prediction blurs to account for the many
   possible futures. In contrast, samples from
   SVG-LP show the digit bouncing off in different plausible
   directions. 
} 
\label{fig:mnist_gen}
\end{figure*}

\begin{figure*}[t!]
\centering
     \includegraphics[width=0.98\linewidth]{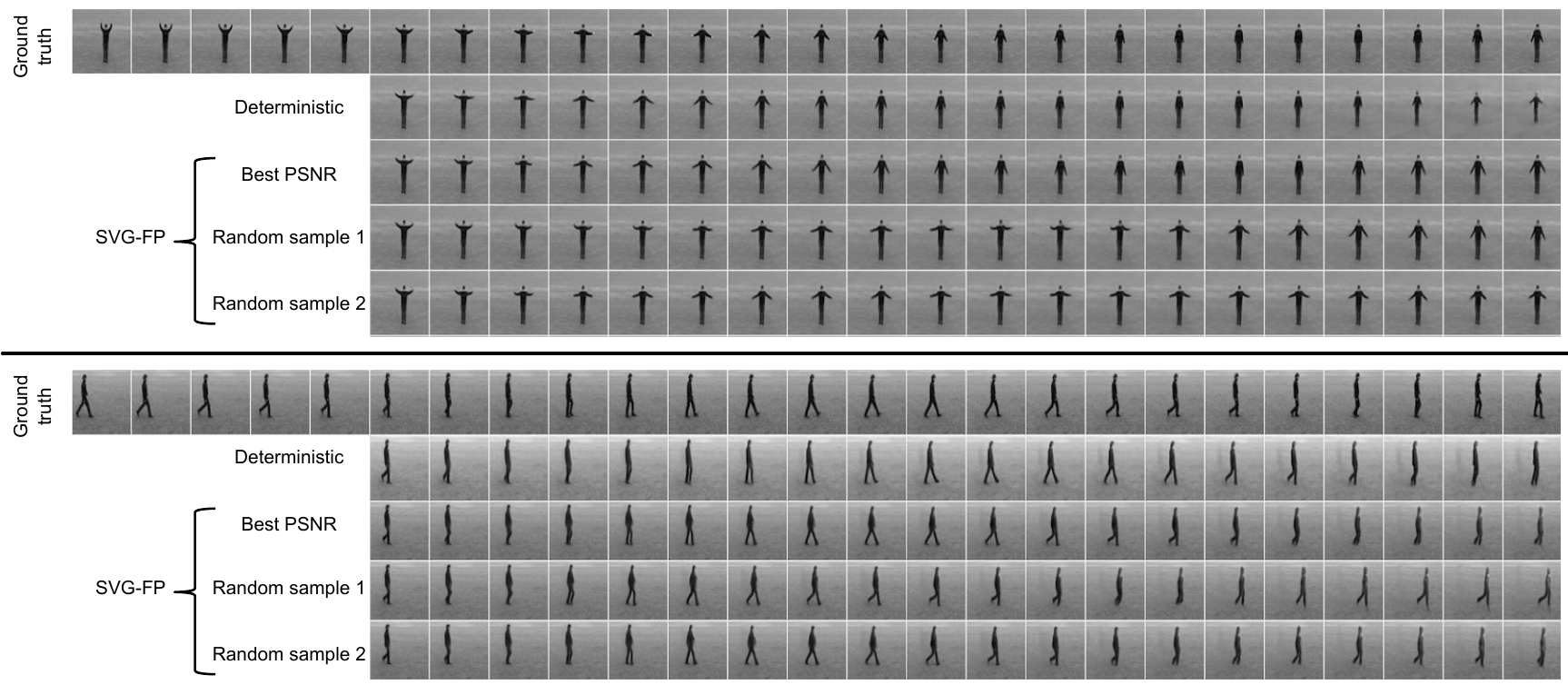}
 \caption{Qualitative comparison between SVG-LP and a purely deterministic baseline. Both models were conditioned on the first 10 frames (the final 5 are shown in the figure) of test sequences. The deterministic model produces plausible predictions for the future frames but frequently mispredicts precise limb locations. In contrast, different samples from SVG-FP reflect the variability on the persons pose in future frames. By picking the sample with the best PSNR, SVG-FP closely matches the ground truth sequence. }
\label{fig:kth_gen}
\end{figure*}

We evaluate our SVG-FP and SVG-LP model on one synthetic video dataset
(Stochastic Moving MNIST) and two real ones (KTH actions \cite{kth} and
BAIR robot \cite{ebert17}).  We show quantitative comparisons by computing
structural similarity (SSIM) and Peak Signal-to-Noise Ratio (PSNR)
scores between ground truth and generated video sequences.  Since
neither of these metrics fully captures perceptual fidelity of
generated sequences we also make a qualitative comparison between
samples from our model and current state-of-the-art methods. 
We encourage the reader to view additional generated videos at: \url{https://sites.google.com/view/svglp/}.

\subsection{Model architectures}
$LSTM_{\theta}$ is a two layer LSTMs with 256 cells in each layer.  
$LSTM_{\phi}$ and $LSTM_{\psi}$ are both single layer LSTMs with 256 cells in each layer.  
Each network has a linear embedding
layer and a fully connected output layer.
The output of $LSTM_{\theta}$ is passed through a tanh nonlinearity before going into the frame decoder.

For Stochastic Moving MNIST, the frame encoder has a DCGAN
discriminator architecture \cite{radford2016} with output dimensionality $|h|$ = 128.
Similarly, the decoder uses a DCGAN generator architecture and a sigmoid output layer.
The output dimensionalities of the LSTM networks are 
$|g| = 128, |\mu_{\phi}| = |\mu_{\psi}| = 10$. 

For KTH and BAIR datasets, the frame encoder uses the same architecture as VGG16 \cite{vgg} up until the final pooling layer with output dimensionality $|h| = 128$. The decoder is a mirrored version of the encoder with pooling layers replaced with spatial up-sampling and a sigmoid output layer.
The output dimensionalities of the LSTM networks are 
$|g| = 128, |\mu_{\phi}| = |\mu_{\psi}| = 32$ for KTH and $|g| = 128, |\mu_{\phi}| = |\mu_{\psi}| = 64$ for BAIR. 

For all datasets we add
skip connections from the encoder at the last ground truth frame to the decoder at $t$, enabling the model to easily generate static
background features.

We also train a deterministic baseline with the same encoder, decoder and LSTM architecture as our frame predictor $p_{\theta}$ but with the latent variables omitted.

\begin{figure*}[t!]
\centering
     \includegraphics[width=0.85\linewidth]{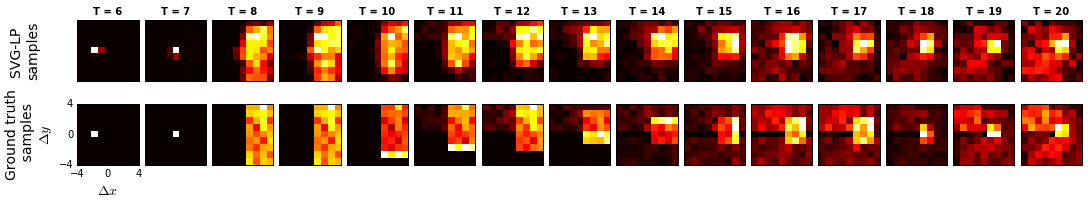}
     \includegraphics[width=0.14\linewidth]{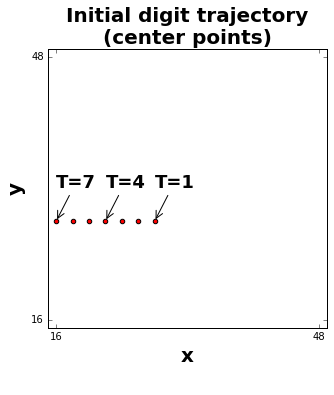}
\noindent\makebox[\linewidth]{\rule{0.99\linewidth}{0.4pt}}
     \includegraphics[width=0.85\linewidth]{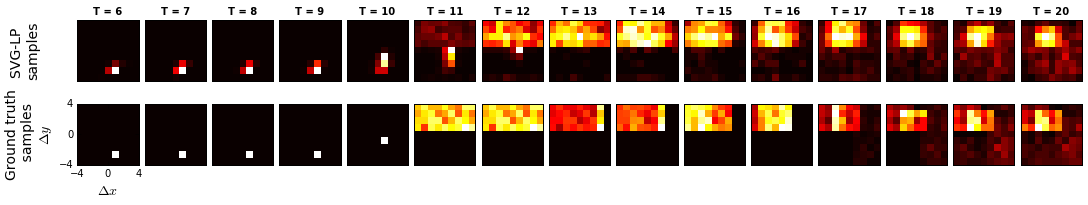}
     \includegraphics[width=0.14\linewidth]{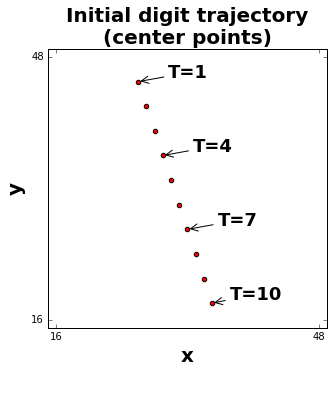}
\noindent\makebox[\linewidth]{\rule{0.99\linewidth}{0.4pt}}
     \includegraphics[width=0.85\linewidth]{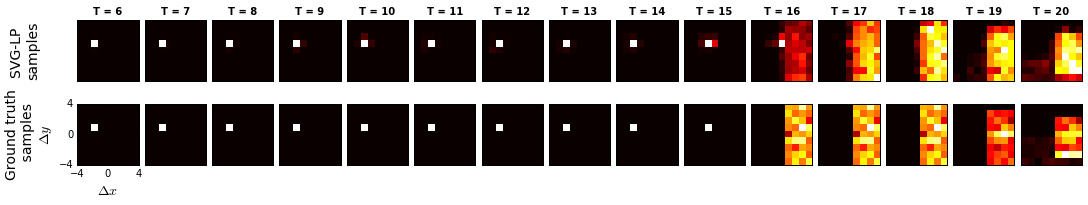}
     \includegraphics[width=0.14\linewidth]{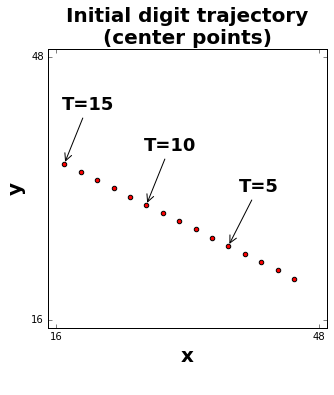}
 \caption{Three examples of our SVG-LP model accurately capturing the
   distribution of MNIST digit trajectories following collision with
   a wall. On the right we show the trajectory of a digit prior to the
   collision. In the ground truth sequence, the angle and speed
   immediately after impact are drawn from at random from uniform distributions.
   Each of
   the sub-plots shows the {\em distribution} of $\Delta x,\Delta y$
   at each time step. In the lower ground truth sequence, the
   trajectory is deterministic before the collision (occurring between
   $t=7$ and $t=8$
   in the first example), corresponding to a delta-function. Following
   the collision, the distribution broadens out to an approximate
   uniform distribution (e.g. $t=8$), before being reshaped by
   subsequent collisions. The upper row shows the distribution
   estimated by our SVG-LP model (after conditioning on ground-truth
   frames from $t=1\ldots5$). Note how our model accurately captures
   the correct distribution many time steps into the future, despite its
   complex shape. The distribution was computed by drawing many
   samples from the model, as well as averaging over different digits
   sharing the same trajectory. The 2nd and 3rd
   examples show different trajectories with correspondingly different impact times
   ($t=11$ and $t=16$ respectively). }
\label{fig:mnist_dist}
\end{figure*}

We train all the models with the ADAM optimizer \cite{adam} and
learning rate $\eta = 0.002$. 
We set $\beta =$ 1e-4 for KTH and BAIR and $\beta =$ 1e-6 for KTH.
Source code and trained models are available at \url{https://github.com/edenton/svg}.

\subsection{Stochastic Moving MNIST}
We introduce the Stochastic Moving MNIST (SM-MNIST) dataset which
consists of sequences of frames of size $64\times64$, containing one
or two MNIST digits moving and bouncing off edge of the frame (walls).
In the original Moving MNIST dataset \cite{Srivastava15} the digits
move with constant velocity and bounce off the walls in a
deterministic manner.  By contrast, SM-MNIST digits move with a
constant velocity along a trajectory until they hit at wall at which
point they bounce off with a random speed and direction.  This
dataset thus contains segments of deterministic motion interspersed
with moments of uncertainty, i.e. each time a digit hits a wall.

Training sequences were
generated on the fly by sampling two different MNIST digits from the
training set (60k total digits) and two distinct trajectories. 
Trajectories were constructed by uniformly sampling $(x, y)$ starting locations and initial velocity vectors $(\Delta x, \Delta y) \in [-4, 4] \times [-4, 4]$.
Every time a digit hits a wall a new velocity vector is sampled.

\begin{figure}[t!]
    \centering
\mbox{
  \includegraphics[width=0.98\linewidth]{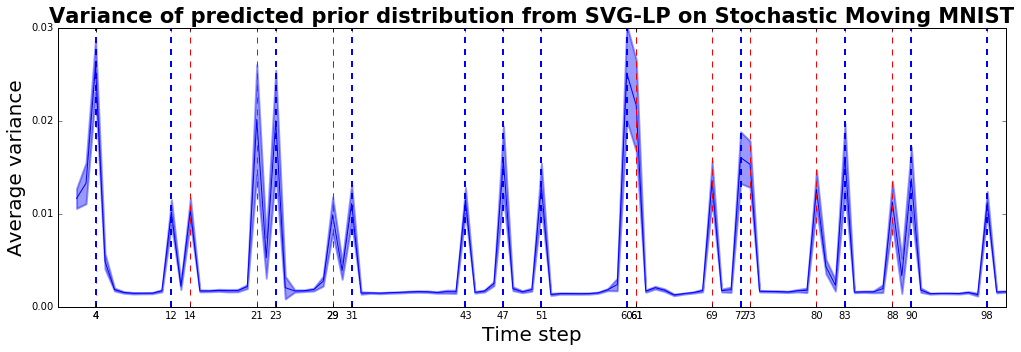}
 }
 \caption{Learned prior of SVG-LP accurately predicts collision points in SM-MNIST. Five hundred test video sequences with different MNIST test digits but synchronized motion were fed into the learned prior. The mean ($\pm$  one standard deviation) of $\sigma_{\psi}(\x_{1:t-1})$ is plotted for $t = 1,...,100$. The true points of uncertainty in the video sequences, i.e. when a digits hits a wall, are marked by vertical lines, colored red and blue for each digit respectively.} 
      \label{fig:mnist_prior}
\vspace{-3mm}
\end{figure}

\begin{figure}[t!]
    \centering
\mbox{
  \includegraphics[width=0.9\linewidth]{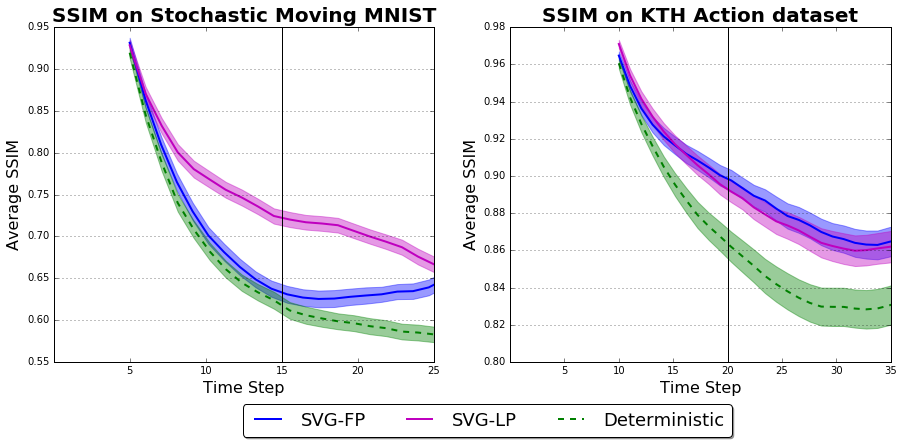}
 }
 \caption{Quantitative evaluation of SVG-FP and SVG-LP video generation quality on
   SM-MNIST (left) and KTH 
   (right). The models are conditioned on the first 5 frames for
   SM-MNIST and 10 frames for KTH. The vertical bar
   indicates the frame number the models were trained to predict up
   to; further generations indicate generalization ability. 
   Mean SSIM over test videos is plotted with 95\% confidence interval shaded. 
     }
    \label{fig:mnist_kth_ssim}
\end{figure}

We trained our SVG models and a deterministic baseline on SM-MNIST by conditioning on 5 frames and training the model to
predict the next 10 frames in the sequence.  
We compute SSIM for SVG-FP and SVG-LP by drawing 100 samples from the
model for each test sequence and picking the one with the best score
with respect to the ground truth. 
\fig{mnist_kth_ssim}(left) plots average SSIM on unseen test videos.
Both SVG-FP and SVG-LP outperform the deterministic baseline and SVG-LP performs best overall, particularly in later time steps.
\fig{mnist_gen} shows sample generations from the deterministic model and SVG-LP. 
Generations from the deterministic model are sharp for several time
steps, but the model rapidly degrades after a digit collides with the wall,
since the subsequent trajectory is uncertain. 

We hypothesize that the improvement of SVG-LP over the SVG-FP model is
due to the mix of deterministic and stochastic movement in the
dataset.  In SVG-FP, the frame predictor must determine how and if the
latent variables for a given time step should be integrated into the
prediction.  In SVG-LP , the burden of predicting points of high
uncertainty can be offloaded to the prior network.  

Empirically, we
measure this in \fig{mnist_prior}. Five hundred different video sequences were
constructed, each with different test digits, but whose
trajectories were synchronized. The plot shows the mean of
$\sigma_{\psi}(\x_{1:t})$, i.e., the variance of the distribution over $\z_t$ predicted by the
learned prior over 100 time steps. Superimposed in red and blue are the time
instants when the the respective digits hit a wall. We see that the learned prior is
able to accurately predict these collisions that result in significant
randomness in the trajectory.


\begin{figure}[t!]
    \centering
\mbox{
  \includegraphics[width=0.9\linewidth]{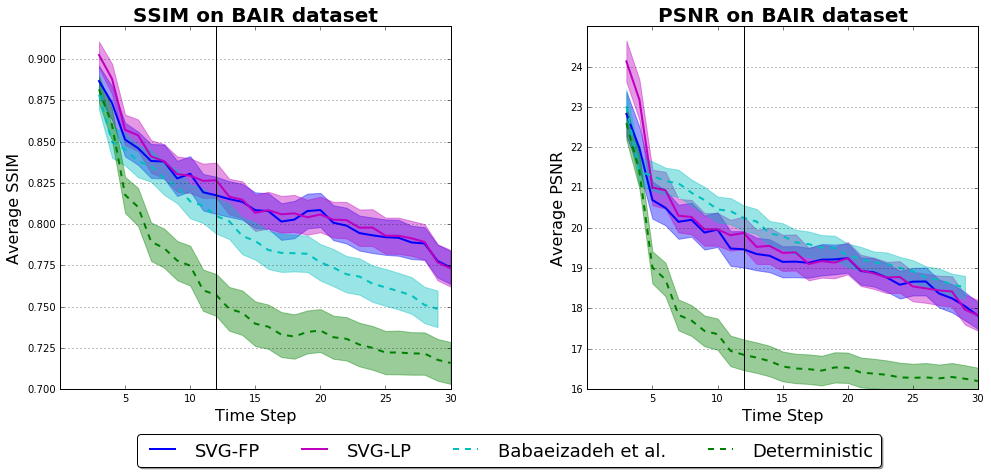}
 }
 \caption{Quantitative comparison between our SVG models and \citet{Babaeizadeh1017} on the BAIR robot dataset. All models are conditioned on the first two frames and generate the subsequent 28 frames. The models were trained to predict up 10 frames in the future, indicated by the vertical bar; further generations indicate generalization ability. Mean SSIM and PSNR over test videos is plotted with 95\% confidence interval shaded.}
    \label{fig:bair_quant}
\end{figure}

\begin{figure}[t!]
    \centering
\mbox{
  \includegraphics[width=0.95\linewidth]{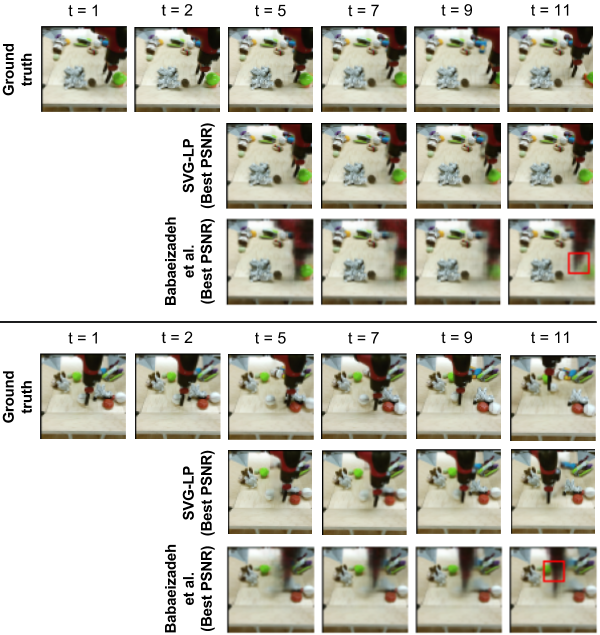}
 }
 \caption{Qualitative comparison between our SVG-LP model and \citet{Babaeizadeh1017}. All models are conditioned on the first two frames of unseen test videos. SVG-LP generates crisper images and predicts plausible movement of the robot arm. }
    \label{fig:bair}
\end{figure}

\begin{figure*}[t!]
    \centering
\mbox{
  \includegraphics[width=\linewidth]{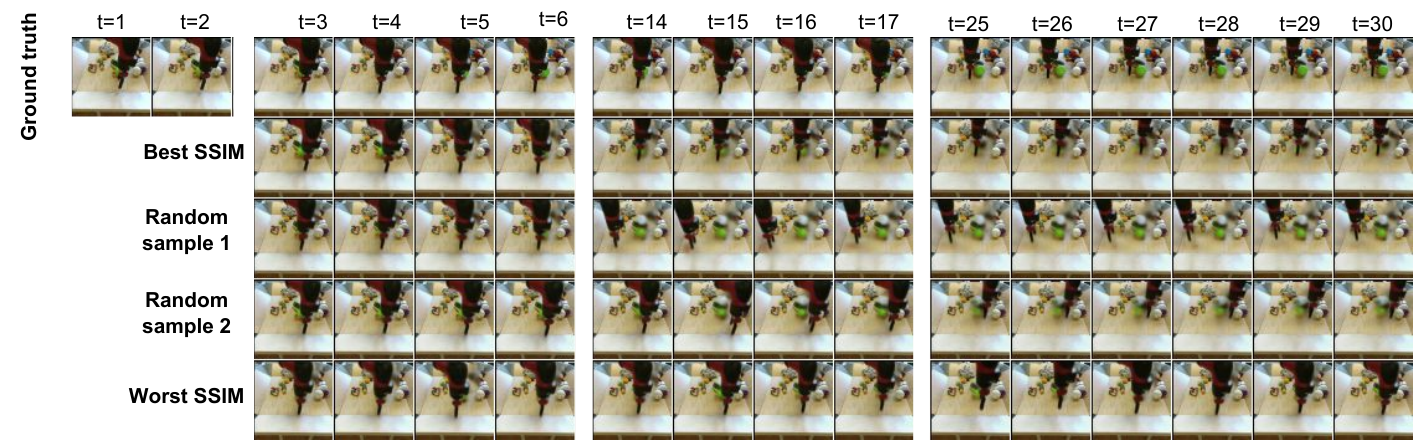}
 }
 \caption{Additional examples of generations from SVG-LP showing crisp and varied predictions. A large segment of the background is occluded in conditioning frames, preventing SVG-LP from directly copying these background pixels into generated frames. In addition to crisp robot arm movement, SVG-LP generates plausible background objects in the space occluded by the robot arm in initial frames. }
    \label{fig:bair_long}
\end{figure*}

\begin{figure*}[t!]
    \centering
\mbox{
  \includegraphics[width=\linewidth]{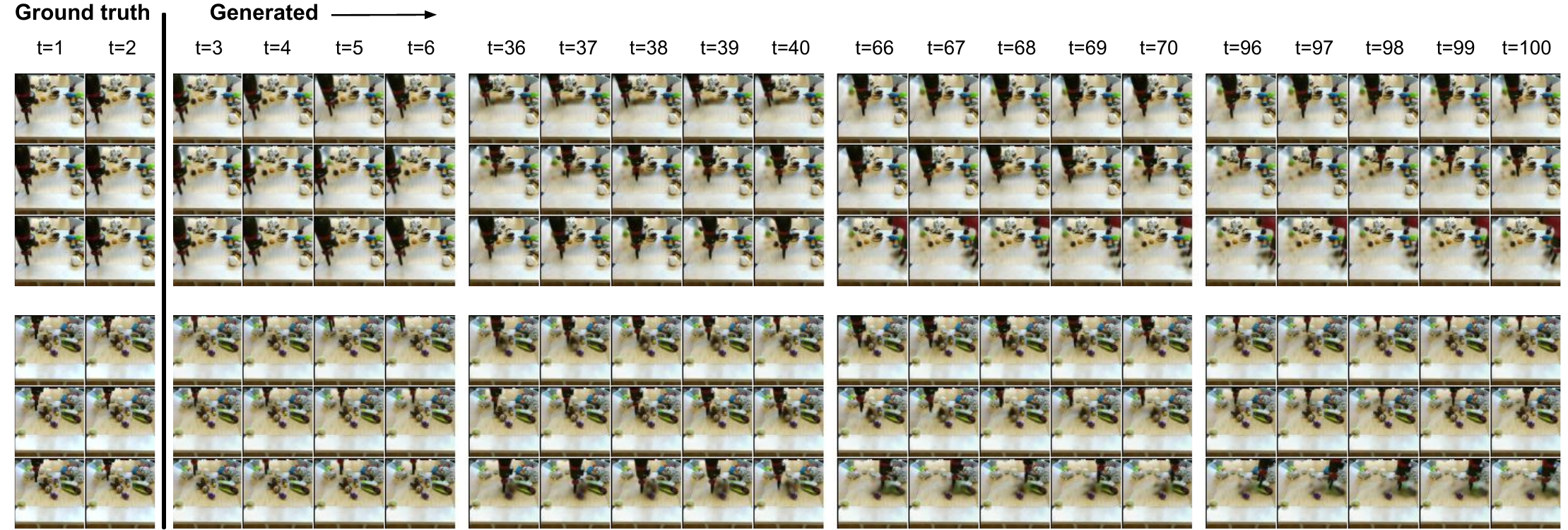}
 }
 \caption{Long range generations from SVG-LP. The robot arm remains crisp up to 100 time steps and object motion can be seen in the generated video frames. Additional videos can be viewed at: \url{https://sites.google.com/view/svglp/}. }
    \label{fig:bair_very_long}
\end{figure*}

One major challenge when evaluating generative video models is
assessing how accurately they capture the full distribution of
possible outcomes, mainly due to the high dimensionality of the space
in which samples are drawn. However, the synthetic nature of single digit SM-MNIST
allows us to investigate this in a principled way. 
A key point to note
is that with each sequence, the digit appearance remains constant with
the only randomness coming from its trajectory once it hits the image
boundary. Thus for a sequence generated from our model, we can
establish the digit trajectory by taking a pair of frames at any time
step and cross-correlating them with the digit used in the initial
conditioning frames. Maxima in each frame reveal the location of the
digit, and the difference between the two gives us the velocity vector
at that time.  By taking an expectation over many samples from our
model (also using the same trajectory but different digits), we can
compute the empirical distribution of trajectories produced by our
model. We can then perform the same operation on a validation set of
ground truth sequences, to produce the true distribution of digit
trajectories and compare it to the one produced by our model.

\fig{mnist_dist} shows SVG-LP (trained on {\em single} digit SM-MNIST) accurately capturing the distribution of MNIST digit trajectories for many time steps. 
The digit trajectory is deterministic before a collision.
This is accurately reflected by the highly peaked distribution of velocity vectors from SVG-LP in the time steps leading up to a collision.
Following a collision, the distribution broadens to approximately uniform before being reshaped by subsequent collisions.
Crucially, SVG-LP accurately captures this complex behavior for many time steps.
The temporally varying nature of the true trajectory distributions further supports the need for a learned prior $p_{\psi}(\z_t | \x_{1:t-1})$.

\subsection{KTH Action Dataset}
The KTH Action dataset \cite{kth} consists of real-world videos of
people performing one of six actions (walking, jogging, running, boxing, handwaving, hand-clapping)
against fairly uniform backgrounds. 
The human motion in the video sequences is fairly regular, however there is still uncertainty regarding the precise locations of the person's joints at subsequent time steps. 
We trained SVG-FP, SVG-LP and the deterministic baseline on $64\times64$ video sequences by conditioning on 10 frames and  training the model to predict the next 10 frames in the sequence.

We compute SSIM for SVG-FP and SVG-LP by drawing 100 samples from the model for each test sequence and picking the one with the best score with respect to the ground truth. 
\fig{mnist_kth_ssim}(right) plots average SSIM on unseen test videos.
SVG-FP and SVG-LP perform comparably on this dataset and both outperform the deterministic baseline.
\fig{kth_gen} shows generations from the deterministic baseline and SVG-FP. 
The deterministic model predicts plausible future frames but, due to the inherent uncertainty in precise limb locations, often deviates from the ground truth.
In contrast, different samples from the stochastic model reflect the variability in future frames indicating the latent variables are being utilized even on this simple dataset. 

\subsection{BAIR robot pushing dataset}
The BAIR robot pushing dataset \cite{ebert17} contains videos of a
Sawyer robotic arm pushing a variety of objects around a table top.
The movements of the arm are highly stochastic, providing a good test
for our model. Although the dataset does contain actions given to the
arm, we discard them during training and make frame
predictions based solely on the video input.

Following \citet{Babaeizadeh1017}, we train SVG-FP, SVG-LP and the deterministic baseline by conditioning on the first two frames of a sequence and predicting the subsequent 10 frames. 
We compute SSIM for SVG-FP and SVG-LP by drawing 100 samples from the model for each test sequence and picking the one with the best score with respect to the ground truth. 
\fig{bair_quant} plots average SSIM and PSNR scores on 256 held out
test sequences, comparing to the state-of-the-art approach of \citet{Babaeizadeh1017}.  
This evaluation consists of conditioning on 2 frames and generating 28
subsequent ones, i.e. longer than at train time,  demonstrating the generalization capability of SVG-FP and SVG-LP. 
Both SVG-FP and SVG-LP outperform \citet{Babaeizadeh1017} in terms of SSIM.
SVG-LP outperforms the remaining models in terms of PSNR for the first few steps, after which  \citet{Babaeizadeh1017} is marginally better. 
Qualitatively, SVG-FP and SVG-LP produce significantly sharper generations than \citet{Babaeizadeh1017}, as illustrated in \fig{bair}.
PSNR is biased towards overly smooth (i.e. blurry) results which might explain the slightly better PSNR scores obtained by \citet{Babaeizadeh1017} for later time steps. 

SVG-FP and SVG-LP produce crisp generations many time steps into the future. \fig{bair_long} shows sample generations up to 30 time steps alongside the ground truth video frames.
We also ran SVG-LP forward for 100 time steps and continue to see
crisp motion of the robot arm (see \fig{bair_very_long}).

\section{Discussion}
We have introduced a novel video prediction model that
combines a deterministic prediction of the next frame with 
stochastic latent variables, drawn from a time-varying distribution learned
from training sequences. 
Our recurrent inference network estimates the latent distribution for each time step allowing easy end-to-end training.
Evaluating the model on real-world sequences, we demonstrate high quality generations that
are comparable to, or better than, existing approaches. On
synthetic data where it is possible to characterize the 
distribution of samples, we see that is able to match complex
distributions of futures.
The framework is sufficiently general that it can readily be
applied to more complex datasets, given appropriate encoder and
decoder modules.

 
\section*{Acknowledgements}
Emily Denton is grateful for the support of a Google PhD fellowship.
This work was supported by ONR \#N00014-13-1-0646 and an NSF CAREER grant.


\bibliography{bibliography}

\begin{thebibliography}{42}
\providecommand{\natexlab}[1]{#1}
\providecommand{\url}[1]{\texttt{#1}}
\expandafter\ifx\csname urlstyle\endcsname\relax
  \providecommand{\doi}[1]{doi: #1}\else
  \providecommand{\doi}{doi: \begingroup \urlstyle{rm}\Url}\fi

\bibitem[Agrawal et~al.(2015)Agrawal, Carreira, and Malik]{agrawal15}
Agrawal, P, Carreira, J, and Malik, J.
\newblock Learning to see by moving.
\newblock In \emph{Proceedings of the International Conference on Computer
  Vision (ICCV)}, 2015.

\bibitem[Babaeizadeh et~al.(2017)Babaeizadeh, Finn, Erhan, Campbell, and
  Levine]{Babaeizadeh1017}
Babaeizadeh, Mohammad, Finn, Chelsea, Erhan, Dumitru, Campbell, Roy~H., and
  Levine, Sergey.
\newblock Stochastic variational video prediction.
\newblock \emph{arXiv:1710.11252}, 2017.

\bibitem[Bayer \& Osendorfer(2014)Bayer and Osendorfer]{bayer2014}
Bayer, Justin and Osendorfer, Christian.
\newblock Learning stochastic recurrent networks.
\newblock \emph{arXiv:1411.7610}, 2014.

\bibitem[Bowman et~al.(2016)Bowman, Vilnis, Vinyals, Dai, Jozefowicz, and
  Bengio]{bowman2016}
Bowman, Samuel~R., Vilnis, Luke, Vinyals, Oriol, Dai, Andrew~M., Jozefowicz,
  Rafal, and Bengio, Samy.
\newblock Generating sentences from a continuous space.
\newblock In \emph{Proceedings of The SIGNLL Conference on Computational
  Natural Language Learning (CoNLL)}, 2016.

\bibitem[Chiappa et~al.(2017)Chiappa, Racaniere, Wierstra, and
  Mohamed]{Chiappa17}
Chiappa, S., Racaniere, S., Wierstra, D., and Mohamed, S.
\newblock Recurrent environment simulators.
\newblock In \emph{Proceedings of the International Conference on Learning
  Representations (ICLR)}, 2017.

\bibitem[Chung et~al.(2015)Chung, Kastner, Dinh, Goel, Courville, and
  Bengio]{chung2015}
Chung, Junyoung, Kastner, Kyle, Dinh, Laurent, Goel, Kratarth, Courville,
  Aaron, and Bengio, Yoshua.
\newblock A recurrent latent variable model for sequential data.
\newblock In \emph{Advances in Neural Information Processing Systems (NIPS)},
  2015.

\bibitem[Denton \& Birodkar(2017)Denton and Birodkar]{denton17}
Denton, Emily and Birodkar, Vighnesh.
\newblock Unsupervised learning of disentangled representations from video.
\newblock In \emph{Advances in Neural Information Processing Systems (NIPS)},
  2017.

\bibitem[Ebert et~al.(2017)Ebert, Finn, Lee, and Levine]{ebert17}
Ebert, Frederik, Finn, Chelsea, Lee, Alex~X., and Levine, Sergey.
\newblock Self-supervised visual planning with temporal skip connections.
\newblock In \emph{Conference on Robot Learning (CoRL)}, 2017.

\bibitem[Finn et~al.(2016)Finn, Goodfellow, and Levine]{Finn16}
Finn, C., Goodfellow, I., and Levine, S.
\newblock Unsupervised learning for physical interaction through video
  prediction.
\newblock In \emph{Advances in Neural Information Processing Systems (NIPS)},
  2016.

\bibitem[Fraccaro et~al.(2016)Fraccaro, Sønderby, Paquet, and
  Winther]{fraccaro2016}
Fraccaro, Marco, Sønderby, Søren~Kaae, Paquet, Ulrich, and Winther, Ole.
\newblock Sequential neural models with stochastic layers.
\newblock In \emph{Advances in Neural Information Processing Systems (NIPS)},
  2016.

\bibitem[Goodfellow et~al.(2014)Goodfellow, Pouget-Abadie, Mirza, Xu,
  Warde-Farley, Ozair, Courville, and Bengio]{goodfellow2014}
Goodfellow, Ian~J., Pouget-Abadie, Jean, Mirza, Mehdi, Xu, Bing, Warde-Farley,
  David, Ozair, Sherjil, Courville, Aaron, and Bengio, Yoshua.
\newblock Generative adversarial nets.
\newblock In \emph{Advances in Neural Information Processing Systems (NIPS)},
  2014.

\bibitem[Goroshin et~al.(2015)Goroshin, Mathieu, and LeCun]{goroshin2015}
Goroshin, Ross, Mathieu, Michael, and LeCun, Yann.
\newblock Learning to linearize under uncertainty.
\newblock In \emph{Advances in Neural Information Processing Systems 28}, 2015.

\bibitem[Henaff et~al.(2017)Henaff, Zhao, and LeCun]{henaff17}
Henaff, Mikael, Zhao, Junbo, and LeCun, Yann.
\newblock Prediction under uncertainty with error-encoding networks.
\newblock \emph{arXiv:1711.04994}, 2017.

\bibitem[Higgins et~al.(2017)Higgins, Matthey, Pal, Burgess, Glorot, Botvinick,
  Mohamed, and Lerchner]{higgins}
Higgins, Irina, Matthey, Loic, Pal, Arka, Burgess, Christopher, Glorot, Xavier,
  Botvinick, Matthew, Mohamed, Shakir, and Lerchner, Alexander.
\newblock Early visual concept learning with unsupervised deep learning.
\newblock In \emph{Proceedings of the International Conference on Learning
  Representations (ICLR)}, 2017.

\bibitem[Jayaraman \& Grauman(2015)Jayaraman and Grauman]{jayaraman15}
Jayaraman, D. and Grauman, K.
\newblock Learning image represen- tations tied to ego-motion.
\newblock In \emph{International Conference on Computer Vision}, 2015.

\bibitem[Kalchbrenner et~al.(2016)Kalchbrenner, van~den Oord, Simonyan,
  Danihelka, Vinyals, Graves, and Kavukcuoglu]{Kalchbrenner16}
Kalchbrenner, N., van~den Oord, A., Simonyan, K., Danihelka, I., Vinyals, O.,
  Graves, A., and Kavukcuoglu, K.
\newblock Video pixel networks.
\newblock \emph{arXiv:1610.00527}, 2016.

\bibitem[Karras et~al.(2017)Karras, Aila, Laine, and Lehtinen]{nvidiagan}
Karras, Tero, Aila, Timo, Laine, Samuli, and Lehtinen, Jaakko.
\newblock Progressive growing of gans for improved quality, stability, and
  variation.
\newblock \emph{arXiv:1710.10196}, 2017.

\bibitem[Kingma \& Ba(2014)Kingma and Ba]{adam}
Kingma, Diederik and Ba, Jimmy.
\newblock Adam: A method for stochastic optimization.
\newblock \emph{Proceedings of the International Conference on Learning
  Representations (ICLR)}, 2014.

\bibitem[Kingma \& Welling(2014)Kingma and Welling]{kingma2014a}
Kingma, D.P. and Welling, M.
\newblock Auto-encoding variational bayes.
\newblock In \emph{Proceedings of the International Conference on Learning
  Representations (ICLR)}, 2014.

\bibitem[Krishnan et~al.(2015)Krishnan, Shalit, and Sontag]{krishnan2015}
Krishnan, R., Shalit, U., and Sontag, D.
\newblock Deep kalman filters.
\newblock \emph{arXiv:1511.05121}, 2015.

\bibitem[Liu(2009)]{liu09}
Liu, C.
\newblock Beyond pixels: exploring new representations and applications for
  motion analysis.
\newblock \emph{PhD thesis, Massachusetts Institute of Technology}, 2009.

\bibitem[Lotter et~al.(2016)Lotter, Kreiman, and Cox]{lotter2016deep}
Lotter, William, Kreiman, Gabriel, and Cox, David.
\newblock Deep predictive coding networks for video prediction and unsupervised
  learning.
\newblock \emph{arXiv:1605.08104}, 2016.

\bibitem[Mathieu et~al.(2016)Mathieu, Couprie, and LeCun]{Mathieu15}
Mathieu, Micha{\"{e}}l, Couprie, Camille, and LeCun, Yann.
\newblock Deep multi-scale video prediction beyond mean square error.
\newblock In \emph{Proceedings of the International Conference on Learning
  Representations (ICLR)}, 2016.

\bibitem[Oh et~al.(2015)Oh, Guo, Lee, Lewis, and Singh]{Oh15}
Oh, J., Guo, X., Lee, H., Lewis, R., and Singh, S.
\newblock Action-conditional video prediction using deep networks in {A}tari
  games.
\newblock In \emph{Advances in Neural Information Processing Systems (NIPS)},
  2015.

\bibitem[Radford et~al.(2016)Radford, Metz, and Chintala]{radford2016}
Radford, Alec, Metz, Luke, and Chintala, Soumith.
\newblock Unsupervised representation learning with deep convolutional
  generative adversarial networks.
\newblock In \emph{Proceedings of the International Conference on Learning
  Representations (ICLR)}, 2016.

\bibitem[Ranzato et~al.(2014)Ranzato, Szlam, Bruna, Mathieu, Collobert, and
  Chopra]{Ranzato14}
Ranzato, Marc'Aurelio, Szlam, Arthur, Bruna, Joan, Mathieu, Micha{\"{e}}l,
  Collobert, Ronan, and Chopra, Sumit.
\newblock Video (language) modeling: a baseline for generative models of
  natural videos.
\newblock \emph{arXiv 1412.6604}, 2014.

\bibitem[Reed et~al.(2017)Reed, van~den Oord, Kalchbrenner, Colmenarejo, Wang,
  Belov, and de~Freitas]{reed17}
Reed, Scott, van~den Oord, Aaron, Kalchbrenner, Nal, Colmenarejo, Sergio~Gomez,
  Wang, Ziyu, Belov, Dan, and de~Freitas, Nando.
\newblock Parallel multiscale autoregressive density estimation.
\newblock In \emph{Proceedings of the International Conference on Machine
  Learning (ICML)}, 2017.

\bibitem[Salimans et~al.(2017)Salimans, Karpathy, Chen, and
  Kingma]{salimans2017pixelcnn++}
Salimans, Tim, Karpathy, Andrej, Chen, Xi, and Kingma, Diederik~P.
\newblock Pixelcnn++: Improving the pixelcnn with discretized logistic mixture
  likelihood and other modifications.
\newblock \emph{arXiv:1701.05517}, 2017.

\bibitem[Schuldt et~al.(2004)Schuldt, Laptev, and Caputo]{kth}
Schuldt, Christian, Laptev, Ivan, and Caputo, Barbara.
\newblock Recognizing human actions: A local svm approach.
\newblock In \emph{Proceedings of the International Conference on Pattern
  Recognition}, 2004.

\bibitem[Simonyan \& Zisserman(2015)Simonyan and Zisserman]{vgg}
Simonyan, K. and Zisserman, A.
\newblock Very deep convolutional networks for large-scale image recognition.
\newblock In \emph{Proceedings of the International Conference on Learning
  Representations (ICLR)}, 2015.

\bibitem[S{\"o}lch et~al.(2016)S{\"o}lch, Bayer, Ludersdorfer, and van~der
  Smagt]{Slch2016VariationalIF}
S{\"o}lch, Maximilian, Bayer, Justin, Ludersdorfer, Marvin, and van~der Smagt,
  Patrick.
\newblock Variational inference for on-line anomaly detection in
  high-dimensional time series.
\newblock \emph{arXiv:1602.07109}, 2016.

\bibitem[Srivastava et~al.(2015)Srivastava, Mansimov, and
  Salakhutdinov]{Srivastava15}
Srivastava, N., Mansimov, E., and Salakhutdinov, R.
\newblock Unsupervised learning of video representations using {LSTM}s.
\newblock In \emph{Proceedings of the International Conference on Machine
  Learning (ICML)}, 2015.

\bibitem[van~den Oord et~al.(2016)van~den Oord, Kalchbrenner, and
  Kavukcuoglu]{Oord16}
van~den Oord, A., Kalchbrenner, N., and Kavukcuoglu, K.
\newblock Pixel recurrent neural networks.
\newblock In \emph{Proceedings of the International Conference on Machine
  Learning (ICML)}, 2016.

\bibitem[Villegas et~al.(2017{\natexlab{a}})Villegas, Yang, Hong, Lin, and
  Lee]{Villegas17a}
Villegas, R., Yang, J., Hong, S., Lin, X., and Lee, H.
\newblock Decomposing motion and content for natural video sequence prediction.
\newblock In \emph{Proceedings of the International Conference on Learning
  Representations (ICLR)}, 2017{\natexlab{a}}.

\bibitem[Villegas et~al.(2017{\natexlab{b}})Villegas, Yang, Zou, Sohn, Lin, and
  Lee]{Villegas17b}
Villegas, Ruben, Yang, Jimei, Zou, Yuliang, Sohn, Sungryull, Lin, Xunyu, and
  Lee, Honglak.
\newblock Learning to generate long-term future via hierarchical prediction.
\newblock In \emph{Proceedings of the International Conference on Machine
  Learning (ICML)}, 2017{\natexlab{b}}.

\bibitem[Vondrick \& Torralba(2017)Vondrick and Torralba]{Vondrick17}
Vondrick, C. and Torralba, A.
\newblock Generating the future with adversarial transformers.
\newblock In \emph{Proceedings of the 2011 IEEE Conference on Computer Vision
  and Pattern Recognition}, 2017.

\bibitem[Vondrick et~al.(2016)Vondrick, Pirsiavash, and Torralba]{Vondrick16}
Vondrick, C., Pirsiavash, H., and Torralba, A.
\newblock Generating videos with scene dynamics.
\newblock In \emph{ar{X}iv 1609.02612}, 2016.

\bibitem[Walker et~al.(2015)Walker, Gupta, and Hebert]{walker15}
Walker, J., Gupta, A., and Hebert, M.
\newblock Dense optical flow prediction from a static image.
\newblock In \emph{ICCV}, 2015.

\bibitem[Wang \& Gupta(2015)Wang and Gupta]{wang2015unsupervised}
Wang, Xiaolong and Gupta, Abhinav.
\newblock Unsupervised learning of visual representations using videos.
\newblock In \emph{CVPR}, pp.\  2794--2802, 2015.

\bibitem[Wiskott \& Sejnowski(2002)Wiskott and Sejnowski]{Wiskott02}
Wiskott, L. and Sejnowski, T.
\newblock Slow feature analysis: Unsupervised learning of invariance.
\newblock \emph{Neural Computation}, 14\penalty0 (4):\penalty0 715--770, 2002.

\bibitem[Xue et~al.(2016)Xue, Wu, Bouman, and Freeman]{visualdynamics16}
Xue, Tianfan, Wu, Jiajun, Bouman, Katherine~L, and Freeman, William~T.
\newblock Visual dynamics: Probabilistic future frame synthesis via cross
  convolutional networks.
\newblock In \emph{Advances in Neural Information Processing Systems (NIPS)},
  2016.

\bibitem[Zou et~al.(2012)Zou, Zhu, Ng, , and Yu.]{Zou12}
Zou, W.~Y., Zhu, S., Ng, A.~Y., , and Yu., K.
\newblock Deep learning of invariant features via simulated fixations in video.
\newblock In \emph{Advances in Neural Information Processing Systems (NIPS)},
  2012.

\end{thebibliography}
\bibliographystyle{icml2018}

\clearpage
\appendix
\section*{Appendix}

\section{Variational bound}
\label{bound}

We first review the variational lower bound on the data likelihood:
\begin{align*}
\log p_{\theta}(\x) &= \log \int_{\z} p_{\theta}(\x | \z) p(\z) \\ 
&= \log \int_{\z} p_{\theta}(\x | \z) p(\z) \frac{q_{\phi}(\z | \x)}{q_{\phi}(\z | \x)}\\
&= \log \mathbb{E}_{q_{\phi}(\z | \x)} \frac{p_{\theta}(\x | \z) p(\z) }{q_{\phi}(\z | \x)}\\
& \ge \mathbb{E}_{q_{\phi}(\z | \x)} \log \frac{p_{\theta}(\x | \z) p(\z) }{q_{\phi}(\z | \x)}\\
&= \mathbb{E}_{q_{\phi}(\z | \x)} \log p_{\theta}(\x | \z) - \mathbb{E}_{q_{\phi}(\z | \x)} \log \frac{q_{\phi}(\z | \x)}{p(\z)}\\
&= \mathbb{E}_{q_{\phi}(\z | \x)} \log p_{\theta}(\x | \z) - D_{KL}(q_{\phi}(\z | \x) || p(\z))\\
\end{align*}

Recall that the SVG frame predictor is parameterized by a recurrent neural network. At each time step the model takes as input $\x_{t-1}$ and $\z_t$ and through the recurrence the model also depends on $\x_{1:t-2}$ and $\z_{1:t-1}$. 
Then, we can further simplify the bound with:
\begin{align*}
\log p_{\theta}(\x | \z) &= \log \prod_t p_{\theta}(\x_t | \x_{1:t-1}, \z_{1:T})\\
&= \sum_t \log p_{\theta}(\x_t | \x_{1:t-1}, \z_{1:t}, \cancel{\z_{t+1:T}})\\
&= \sum_t \log p_{\theta}(\x_t | \x_{1:t-1}, \z_{1:t})
\end{align*}

Recall, the inference network used by SVG-FP and SVG-LP is parameterized by a recurrent neural network that outputs a different distribution $q_{\phi}(\z_t | \x_{1:t})$ for every time step $t$. 
Let $\z = [\z_1, ..., \z_T]$ denote the collection of latent variables across all time steps and $q_{\phi}(\z | \x)$ denote the distribution over $\z$. Due to the independence across time, we have $$q_{\phi}(\z | \x) = \prod_t q_{\phi}(\z_t | \x_{1:t})$$
The independence of $\z_1, ..., \z_T$ allows the $D_{KL}$ term of the loss to be decomposed into individual time steps:
\begin{align*}
&D_{KL}(q_{\phi}(\z | \x) || p(\z)) \\
&= \int_\z q_{\phi}(\z | \x) \log \frac{q_{\phi}(\z | \x)}{p(\z)}\\
&=\int_{\z_1} ...  \int_{\z_T} q_{\phi}(\z_1 | \x_1)...q_{\phi}(\z_T | \x_{1:T}) \log \frac{q_{\phi}(\z_1 | \x_1)...q_{\phi}(\z_t | \x_{1:T})}{p(\z_1)...p(\z_T)}\\
&=\int_{\z_1}  ...  \int_{\z_T} q_{\phi}(\z_1 | \x_1)...q_{\phi}(\z_T | \x_{1:T}) \sum_t \log \frac{q_{\phi}(\z_t | \x_{1:t})}{p(\z_t)} \\
&=\sum_t \int_{\z_1}  ...  \int_{\z_T} q_{\phi}(\z_1 | \x_1)...q_{\phi}(\z_T | \x_{1:T}) \log \frac{q_{\phi}(\z_t | \x_{1:t})}{p(\z_t)} \\
&\text{And because } \int_x p(x) = 1 \text{  this simplifies to:}\\
&= \sum_t \int_{\z_t} q_{\phi}(\z_t | \x_{1:t}) \log \frac{q_{\phi}(\z_t | \x_{1:t})}{p(\z_t)}\\
&= \sum_t D_{KL}(q_{\phi}(\z_t | \x_{1:t}) || p(\z_t))\\
\end{align*}

Putting this all together we have:
\begin{align*}
\log p_{\theta}(\x) &\geq \mathcal{L}_{\theta, \phi}(\x_{1:T}) \\
&=  \mathbb{E}_{q_{\phi}(\z | \x)} \log p_{\theta}(\x | \z) -  D_{KL}(q_{\phi}(\z | \x) || p(\z)) \\
&=  \sum_t \big[ \mathbb{E}_{q_{\phi}(\z_{1:t} | \x_{1:t})}   \log p_{\theta}(\x_t | \x_{1:t-1}, \z_{1:t}) \\
& \hspace{13mm} -   D_{KL}(q_{\phi}(\z_t | \x_{1:t}) || p(\z_t)) \big] \\
\end{align*}

\small

\end{document}